\documentclass[10pt,letterpaper,journal]{IEEEtran}
\usepackage[T1]{fontenc}
\usepackage[utf8]{inputenc}
\usepackage{amsmath,amssymb}
\usepackage{booktabs,multirow,array,tabularx}
\usepackage{graphicx}
\usepackage{algorithm}
\usepackage[noend]{algpseudocode}
\usepackage{cite}
\usepackage{url}
\usepackage{stfloats}
\usepackage{placeins}
\usepackage[hidelinks]{hyperref}
\graphicspath{{figure/}}

\algrenewcommand\algorithmicrequire{\textbf{Input:}}
\algrenewcommand\algorithmicensure{\textbf{Output:}}
\begin{document}
\title{A Multi-Parent Edge Assembly Crossover Driven Hybrid Genetic Algorithm for the Hamiltonian \textit{p}-Median Problem}

\author{Junkai Huang, Pengfei He, Jin-Kao Hao, and Xin Yao%
\thanks{Junkai Huang is with the School of Automation and Key Laboratory of Measurement and Control of Complex Systems of Engineering, Ministry of Education, Southeast University, Nanjing 210096, China (e-mail: zengke.janx@gamil.com).}%
\thanks{Pengfei He is with the School of Data Science, Lingnan University, Hong Kong SAR, China, and also with the School of Automation and Key Laboratory of Measurement and Control of Complex Systems of Engineering, Ministry of Education, Southeast University, Nanjing 210096, China (e-mail: pengfeihe606@gmail.com). Corresponding author: Pengfei He.}%
\thanks{Jin-Kao Hao is with LERIA, Universit\'{e} d'Angers, 2 Boulevard Lavoisier, 49045 Angers, France (e-mail: jin-kao.hao@univ-angers.fr).}%
\thanks{Xin Yao is with the School of Data Science, Lingnan University, Hong Kong SAR, China (e-mail: xinyao@ln.edu.hk).}}
\maketitle
\begin{abstract}
The Hamiltonian \textit{p}-median problem seeks to partition the vertices of a graph into exactly $p$ disjoint cycles while minimizing their total cost. This paper presents a hybrid genetic algorithm incorporating a multi-parent edge assembly crossover operator. The crossover combines a base solution with multiple donor solutions in a single recombination operation, thereby expanding the space of available edge exchanges. It uses virtual copies of the base solution to establish edge balance, together with an auxiliary directed graph and a budgeted search to construct valid AB-cycles. Extensive experiments on 215 benchmark instances across four sets show that the proposed algorithm improves the best-known solution values for 34 instances and matches those for another 168. Additional experiments indicate that a moderate number of parents provides beneficial diversity, whereas excessive parent counts, particularly when $p$ is large, lead to excessive population dispersion and hinder convergence.
\end{abstract}
\begin{IEEEkeywords}
Hamiltonian \textit{p}-median problem, edge assembly crossover, multi-parent crossover, hybrid genetic algorithm.
\end{IEEEkeywords}

\section{Introduction}
\label{sec:introduction}

The Hamiltonian \textit{p}-median problem (HpMP) is defined on a complete edge-weighted graph and consists of finding exactly $p$ vertex-disjoint cycles of minimum total cost such that every vertex belongs to one cycle. It combines partitioning decisions with cyclic routing decisions and is relevant to school and depot location \cite{canas2025location}. Other application contexts, including industrial process scheduling, laser multi-scanning, and leather cutting, are discussed in Herr{\'a}n et~al.~\cite{HerranColmenarDuarte2019} and He et~al.~\cite{HeHaoWu2024}. The symmetric traveling salesman problem (TSP) is the particular case with $p=1$. Consequently, the HpMP is also a $\mathcal{NP}$-hard problem and remains computationally challenging for large instances \cite{BrancoCoelho1990,canas2025location}.

Due to its theoretical interest and practical relevance, the HpMP has motivated many polyhedral investigations \cite{BarbatoGouveia2024}, exact algorithms \cite{ErdoganLaporteRodriguezChia2016,BektasGouveiaSantos2019,BarbatoGouveia2024,canas2025location} and dedicated heuristic methods \cite{ErdoganLaporteRodriguezChia2016,HerranColmenarDuarte2019,HeHaoWu2024}. Although recent exact algorithms have solved selected benchmark instances with up to 400 vertices, many instances remain unsolved within the prescribed time limits \cite{BarbatoGouveia2024,canas2025location}. Metaheuristic algorithms therefore remain important for obtaining high-quality solutions to larger or computationally difficult instances. He et~al.~\cite{HeHaoWu2024} proposed a state-of-the-art hybrid genetic algorithm (HGA) that effectively solves the HpMP by combining an edge assembly crossover (EAX) operator, a local search procedure employing multiple neighborhood structures, and an advanced population management strategy. Indeed, EAX is well suited to the HpMP, as the problem generalizes the traveling salesman problem (TSP), for which EAX was originally designed \cite{NagataKobayashi2013}.

The EAX operator employed in HGA uses two parents to generate an offspring solution \cite{HeHaoWu2024}. Offspring generated by EAX mainly inherit edges from the two selected parents. When these parents share many edges, the remaining parental edges offer limited alternatives for recombination \cite{HeHaoWu2024,he2022general}. This can restrict offspring diversity and increase the risk of premature convergence, even when other population members contain complementary edges.

Multi-parent recombination has been studied as an extension of conventional two-parent crossover \cite{EibenRaueRuttkay1994}. From a search perspective, multi-parent crossover can also be viewed as a way of extending the recombination neighborhood of a base solution, by allowing offspring to draw on complementary information from several parents. This perspective shares the motivation of extended-neighborhood search in simulated annealing \cite{Yao1991Extended,Yao1992Dynamic} and evolutionary optimization \cite{LiangYaoNewtonHoffman2002,TangMeiYao2009}. For the TSP, Sakai et~al.~\cite{SakaiHanadaOrito2018} proposed an EAX variant that uses more than two parents in a single crossover to increase offspring diversity. For instance, Porumbel et~al.~\cite{PorumbelHaoKuntz2009} proposed a multi-parent crossover operator for the graph coloring problem, in which backbone information extracted from the parent solutions is used to construct offspring solutions. More recently, Zou et~al.~\cite{ZouHaoWu2024} introduced an EAX with triple parents for the latency location routing problem, where an EAX is first applied to two parents, and the resulting offspring solutions are then recombined with the third parent through another EAX to produce final offspring solutions. However, since this operator applies EAX sequentially rather than combining multiple parents simultaneously, it does not fully exploit the potential of multi-parent crossover.

To fill this gap and propose more powerful crossover operators, we propose a multi-parent edge assembly crossover operator (mpEAX), in which one parent serves as the base solution and the remaining parents act jointly as donors within a single recombination operation. The selected parents contribute jointly to a single recombination, rather than through successive binary crossovers. The main contributions of this work are summarized as follows.
\begin{itemize}
\item We propose mpEAX, a multi-parent extension of EAX that combines complementary edge information from one base solution and several donor solutions in a single crossover. The operator is integrated into the existing HGA framework to form $\mathrm{HGA}_{mp}$.

\item We analyze how the parent count affects search behavior and performance, clarifying the benefits and limitations of multi-parent recombination.

\item The proposed algorithm improves 34 best-known results and matches 168 others out of 215 benchmark instances. Additional experiments clarify the influence of the number of parent solutions. Moreover, the publicly available source code of the algorithm can facilitate the practical solution of real-life applications that can be formulated as the HpMP.
\end{itemize}

The remainder of the paper is organized as follows. Section~\ref{sec:related} defines the HpMP and reviews EAX and existing multi-parent studies. Section~\ref{sec:mpeax} presents the proposed mpEAX operator. Section~\ref{sec:experimental-evaluation} reports a benchmark comparison against published results. Section~\ref{sec:parent-effect} examines the effect of the parent count and explains why an excessive number of parents degrades performance on instances. Section~\ref{sec:conclusion} concludes the paper.

\section{Background and related work}
\label{sec:related}
In this section, we briefly review the HpMP, the EAX operator, and existing multi-parent crossover operators.

\subsection{Algorithms for the HpMP}
The HpMP was originally introduced as a joint clustering and cyclic routing problem \cite{BrancoCoelho1990}. Glaab and Pott~\cite{GlaabPott2000} investigated its combinatorial structure, identified tractable special cases on dedicated graph structures, and established its complexity properties. Subsequent research has progressively strengthened exact mathematical formulations: Erdo{\u{g}}an et~al.~\cite{ErdoganLaporteRodriguezChia2016} proposed a branch-and-cut algorithm based on an enhanced $p$-median formulation, Bekta{\c{s}} et~al.~\cite{BektasGouveiaSantos2019} developed a directed formulation and devised a branch-and-cut algorithm, and Barbato and Gouveia~\cite{BarbatoGouveia2024} advanced polyhedral studies with corresponding branch-and-cut procedures. More recently, Canas and Gouveia~\cite{canas2025location} introduced novel compact formulations with fewer variables, along with branch-and-cut algorithms capable of effectively solving instances with up to 400 vertices. Nevertheless, these approaches remain limited in their ability to handle large instances.

Consequently, metaheuristic algorithms have become the method of choice for instances that exceed the capabilities of exact methods. Erdo{\u{g}}an et~al.~\cite{ErdoganLaporteRodriguezChia2016} proposed an iterated local search that integrates a giant-tour representation with dynamic programming and routing moves. Herr{\'a}n et~al.~\cite{HerranColmenarDuarte2019} developed a general variable neighborhood search featuring five routing-based neighborhood operators, multi-start variants, and parallel migration strategies. Both frameworks adopt a trajectory-based paradigm, where diversification is driven by neighborhood modifications along one or more search paths. In contrast, He et~al.~\cite{HeHaoWu2024} introduced a population-based hybrid genetic algorithm that combines EAX, multi-neighborhood local search, mutation, and population management, and has demonstrated strong scalability on large instances.

\subsection{EAX and EAX-based hybrid search}

The EAX operator was introduced by Nagata and Kobayashi~\cite{NagataKobayashi1997} for the TSP and has since been developed into a powerful genetic algorithm \cite{Nagata2006,NagataKobayashi2013}. He et~al.~\cite{HeHaoWu2024} then adapted EAX to solve the HpMP. Consider two feasible parent solutions $\varphi_A$ and $\varphi_B$ with edge sets $\mathcal{E}_{\varphi_A}$ and $\mathcal{E}_{\varphi_B}$, respectively. First, common edges from both parents are removed, and non-common edges are used to construct the joint graph $\mathcal{G}_{AB} = \bigl(\mathcal{V},\ \mathcal{E}_{\varphi_A} \triangle \mathcal{E}_{\varphi_B}\bigr)$ where $\mathcal{E}_{\varphi_A} \triangle \mathcal{E}_{\varphi_B} = \bigl(\mathcal{E}_{\varphi_A} \setminus \mathcal{E}_{\varphi_B}\bigr) \cup \bigl(\mathcal{E}_{\varphi_B} \setminus \mathcal{E}_{\varphi_A}\bigr)$ denotes the symmetric difference of the two edge sets. The joint graph $\mathcal{G}_{AB}$ defines the exchange space , i.e., the set of candidate edge exchanges supported by this graph: the common edges are already fixed, while every remaining edge is available for alternating replacement.

Because every vertex is associated with degree of two in each parent solution (in both the TSP and the HpMP), the degree of each vertex $v$ in the joint graph $\mathcal{G}_{AB}$ is necessarily balanced where $\deg_{\mathcal{E}_{\varphi_A}\setminus\mathcal{E}_{\varphi_B}}(v) = 2 - \deg_{\mathcal{E}_{\varphi_A}\cap\mathcal{E}_{\varphi_B}}(v) = \deg_{\mathcal{E}_{\varphi_B}\setminus\mathcal{E}_{\varphi_A}}(v)$. Hence, at every vertex $v \in \mathcal{V}$, the edges in $\mathcal{G}_{AB}$ appear in pairs belonging alternately to $\mathcal{E}_{\varphi_A}$ and $\mathcal{E}_{\varphi_B}$. This allows us to construct an edge sequence by alternately selecting edges from the two parents. Traversing the joint graph according to this alternating rule eventually returns to the starting vertex after an even number of steps, forming an \textit{AB-cycle}. Iterating this process over the remaining unused edges partitions $\mathcal{G}_{AB}$ into \textit{AB-cycles} that cover all edges.

The resulting \textit{AB-cycles} are grouped into \textit{E-sets}, each comprising one or more cycles, to generate offspring through exchanges of parental edges. Given a selected \textit{E-set}, an intermediate solution is constructed from a base solution, say $\varphi_A$, by removing the edges of $\mathcal{E}_A$ contained in the \textit{E-set} and inserting those of $\mathcal{E}_B$ contained in it, while retaining all other edges of $\mathcal{E}_A$. Although this exchange preserves degree two at every vertex, it may produce disconnected subtours. A repair procedure, typically based on the 2-opt* operator, is therefore applied to restore feasibility. A repair procedure based on the 2-opt* operator is then applied to restore exactly $p$ cycles \cite{HeHaoWu2024}.

The effect of \textit{E-set} size on offspring quality is not necessarily monotonic: smaller \textit{E-sets} favor local intensification around the base solution, whereas larger ones permit more substantial structural changes but may produce more subtours and incur greater repair costs. Different EAX variants \cite{Nagata2006} balance this trade-off in different ways, and Heins et~al.~\cite{HeinsWhitleyKerschke2025} further show that \textit{AB-cycle} selection and the associated repair overhead are closely linked to the fitness of offspring solutions. This trade-off is particularly important for the HpMP, where an edge exchange can affect both the internal connectivity of individual routes and the total number of routes \cite{HeHaoWu2024}. Consequently, careful \textit{AB-cycle} selection and control of \textit{E-set} size are essential during offspring generation \cite{HeHaoWu2024}.

High-quality local optima often share many edges \cite{he2022general,HeHaoWu2024}. Although this shared structure is valuable, it can restrict recombination by EAX: common edges do not participate in the exchange, so \textit{AB-cycles} must be formed from the non-common edges of the selected pair. When the two parents are highly similar, few edges remain available for constructing exchanges, even though other high-quality solutions in the population may contain complementary edges. This observation motivates the use of multiple parent solutions within a single crossover.

\subsection{Multi-parent crossover operators}

Multi-parent crossover operators have been studied since the early days of evolutionary computation. Eiben et~al.~\cite{EibenRaueRuttkay1994} demonstrated that their effectiveness is problem-dependent, with performance gains often diminishing or varying non-monotonically as the number of parents increases. A recurring principle is that effective multi-parent crossover relies on preserving problem-relevant structural information. Examples include inheriting high-quality color classes in graph coloring \cite{PorumbelHaoKuntz2009}, preserving a common partition backbone \cite{BenlicHao2011}, extending permutation mappings \cite{TingSuLee2010}, and eliminating overlapping sequences in scheduling \cite{ChenChenChangMani2013}. Wang et~al.~\cite{WangLuHao2010} further examined the behavioral differences among multi-parent crossover operators through the population structures they induce and the quality of the offspring they generate. Although these studies highlight the importance of problem-specific crossover design, their grouping-, bit-string-, and permutation-based mechanisms lack the balanced alternation of parental edges that underpins EAX.

The closest precedent in the EAX literature is the work of Sakai et~al.~\cite{SakaiHanadaOrito2018}, who incorporated more than two parents into a single TSP recombination operation, allowing edges from multiple parents to contribute jointly to {an offspring solution}. This approach differs from that of Zou et~al.~\cite{ZouHaoWu2024}, who combine three parents through two sequential binary crossover operations: two parents are first recombined, and the resulting offspring is then recombined with the third parent before repairing solutions.

Extending EAX to multiple parents requires more than pooling parental edges. Combining one base solution with $k-1$ donors disrupts the balance between base and donor edges at each vertex, which is required to construct alternating \textit{AB-cycles}. Moreover, the same undirected edge may appear in multiple donors, while the expanded edge pool may yield excessively long cycles. A multi-parent crossover must therefore maintain the required edge balance, prevent conflicting operations on the same edge, and control the size of the resulting exchanges.

To address these challenges, we propose mpEAX, which combines one base solution and $k-1$ donor solutions in a balanced joint multigraph, constructs \textit{AB-cycles}, and maps compatible cycles to executable edge exchanges. This design enables a single recombination operator to draw on complementary edge information from multiple parents while preserving the edge-based recombination principle of EAX. In particular, an \textit{AB-cycle} can combine donor edges from different parents, allowing joint exchanges that cannot be formed from the base solution and any single donor alone. Thus, mpEAX provides a problem-specific way of extending the recombination neighborhood. As with extended-neighborhood search, the aim is to explore additional promising exchanges rather than simply maximize the magnitude of structural changes \cite{TangMeiYao2009}.

\section{Multi-parent edge assembly crossover}
\label{sec:mpeax}

Let $\Phi=\{\varphi_A,\varphi_{B_1},\ldots,\varphi_{B_{k-1}}\}$ denote the selected family of $k$ parents, where $\varphi_A$ is the base solution and $\Phi_B=\{\varphi_{B_1},\ldots,\varphi_{B_{k-1}}\}$ contains the $k-1$ donor solutions. If donor edges are pooled directly with those of $\varphi_A$, each vertex is incident to two base edges but $2(k-1)$ donor edges, counting multiplicities. This creates a vertex-wise imbalance between base and donor edges, violating the balance required for \textit{AB-cycle} construction. To restore this balance, mpEAX introduces $k-2$ virtual copies of the base solution and forms $\Phi_A=\{\varphi_A,\varphi_{A_1},\ldots,\varphi_{A_{k-2}}\}$. The two balanced parent families then serve as the input for constructing the joint multigraph $\mathcal{G}_{mp}$.

\subsection{Balanced joint multigraph}
\label{sec:balance}

Let $\mathcal{E}_{\Phi_A}=\bigcup_{\varphi\in \Phi_A}\mathcal{E}_\varphi$ and $\mathcal{E}_{\Phi_B}=\bigcup_{\varphi\in \Phi_B}\mathcal{E}_\varphi$ denote the edge multisets of $\Phi_A$ and $\Phi_B$, respectively, where $|\mathcal{E}_{\Phi_A}| = |\mathcal{E}_{\Phi_B}|$. Because $\mathcal{G}_{\Phi_A}$ and $\mathcal{G}_{\Phi_B}$ are multigraphs, parallel edges may arise within each multiset. For edge $e$, let $n^a_e$ and $n^b_e$ denote its multiplicities in $\mathcal{E}_{\Phi_A}$ and $\mathcal{E}_{\Phi_B}$, respectively. Common parallel edges are eliminated by removing $\min(n^a_e,n^b_e)$ copies of $e$ from both multisets, yielding reduced edge multisets $\mathcal{E}^{r}_{\Phi_A}$ and $\mathcal{E}^{r}_{\Phi_B}$ with corresponding reduced graphs $\mathcal{G}^{r}_{\Phi_A}$ and $\mathcal{G}^{r}_{\Phi_B}$. The joint multigraph is then defined as $\mathcal{G}_{mp} = (\mathcal{V},\mathcal{E})$, where $\mathcal{E} = \mathcal{E}^{r}_{\Phi_A}\cup\mathcal{E}^{r}_{\Phi_B}$. By construction, every vertex $v\in\mathcal{V}$ satisfies the balance condition $\deg_{\mathcal{E}_{\Phi_A}^r}(v)=\deg_{\mathcal{E}_{\Phi_B}^r}(v)$.

\begin{figure*}[!t]
\centering
\includegraphics[width=0.8\textwidth]{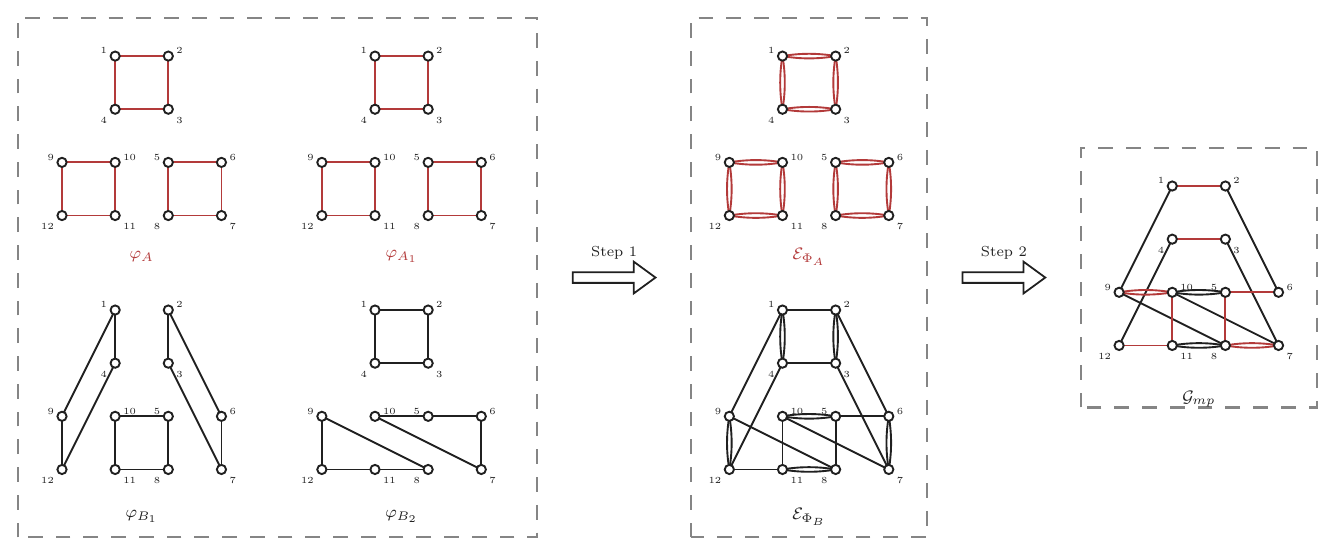}
\caption{The illustration of joint multigraph $\mathcal{G}_{mp}$.}
\label{fig:mpeax_first_step}
\end{figure*}

{Fig.~\ref{fig:mpeax_first_step} shows an illustrative example of creating $\mathcal{G}_{mp}$, where} three parent solutions ($\varphi_A$, $\varphi_{B_1}$, and $\varphi_{B_2}$) are involved to generate offspring solutions, the virtual copy $\varphi_{A_1}$ is added. Then, the joint multigraph $\mathcal{G}_{mp}$ is constructed where common parallel edges are removed. For example, parallel edges (1,4) are removed from both $\mathcal{E}_{\Phi_A}$ and $\mathcal{E}_{\Phi_B}$.

\subsection{Auxiliary directed graph}
\label{sec:auxiliary-graph}

In the joint multigraph $\mathcal{G}_{mp}$, multiple parallel edges may represent the same undirected edge $(u,v)$, such as $(9,10)$ in Fig.~\ref{fig:mpeax_first_step}. Since EAX constructs \textit{AB-cycles} by alternating between base and donor edges \cite{HeHaoWu2024}, treating all parallel edges as distinct alternatives may produce cycles that repeatedly traverse the same underlying edge. Such repetitions introduce redundancy and limit the incorporation of distinct parental edges into offspring. For example, starting from vertex~5 in Fig.~\ref{fig:mpeax_first_step}, alternating edge selection may yield the sequence $(5,10,11,8,7,10,9,8,7,3,4,12,11,8,5)$, in which the underlying edges $(8,11)$ and $(7,8)$ are each traversed twice.

Treating parallel edges as mutually exclusive alternatives, however, creates another difficulty. Under this restriction, each underlying undirected edge can be traversed at most once within an \textit{AB-cycle}: selecting one copy makes all remaining edges inadmissible within that cycle. This restriction can disrupt the symmetric difference of the joint multigraph. In the above example, selecting $(8,7)$ leaves vertex~7 with a residual degree of~3 after the remaining edges become inadmissible. Consequently, the edge multiplicities represented in $\mathcal{G}_{mp}$ no longer reflect the number of permissible edge operations, complicating the direct construction of valid \textit{AB-cycles}. To address this problem, we transform the joint multigraph into an auxiliary directed graph.

To construct \textit{AB-cycles} efficiently, we aggregate all parallel edges joining the same pair of vertices into a single edge group. Since an edge group is undirected and thus does not specify a traversal direction, each group is represented by two directional states, which form the nodes of an auxiliary directed graph. Specifically, the joint multigraph $\mathcal{G}_{mp}=(\mathcal{V},\mathcal{E})$ is transformed into an auxiliary graph $\mathcal{G}'=(\mathcal{S},\mathcal{A})$. All parallel edges joining vertices $u$ and $v$ in $\mathcal{G}_{mp}$ are mapped to two state nodes, $s^{uv}$ and $s^{vu}$, representing traversal from $u$ to $v$ and from $v$ to $u$, respectively. Consequently, $|\mathcal{S}|$ is twice the number of distinct vertex pairs connected by edges in $\mathcal{G}_{mp}$, regardless of their multiplicities. Each state node $s^{uv}\in\mathcal{S}$ is endowed with a parameter $c_{s^{uv}}\in\{0,1\}$ which indicates whether the associated edge group belongs to $\mathcal{E}_{\Phi_A}^{r}$ or $\mathcal{E}_{\Phi_B}^{r}$.

The arc set $\mathcal{A}$ specifies the admissible transitions between directional states. An arc links $s^{uv}$ to $s^{vw}$ if the two states meet at vertex $v$, represent edges from opposite sides of the joint multigraph, and do not immediately traverse the same undirected edge in reverse. Formally, $\mathcal{A} = \left\{(s^{uv},s^{vw})\in\mathcal{S}\times\mathcal{S}| u,v,w\in\mathcal{V},u\neq w,\; c_{s^{uv}}\neq c_{s^{vw}} \right\}$.

A state node $s^{uv}$ represents traversal of the edge group joining $u$ and $v$ from $u$ to $v$. An arc from $s^{uv}$ to $s^{vw}$ continues the walk through $v$ using an edge group from the opposite side of the joint multigraph, without immediately reversing along the same undirected edge. Thus, a directed path in $\mathcal{G}'$ corresponds to an alternating edge walk in $\mathcal{G}_{mp}$. A closed path induces a legal \textit{AB-cycle} if each undirected edge group appears at most once, excluding the repeated start state used to indicate closure.

\begin{figure*}[!t]
\centering
\includegraphics[width=0.8\textwidth]{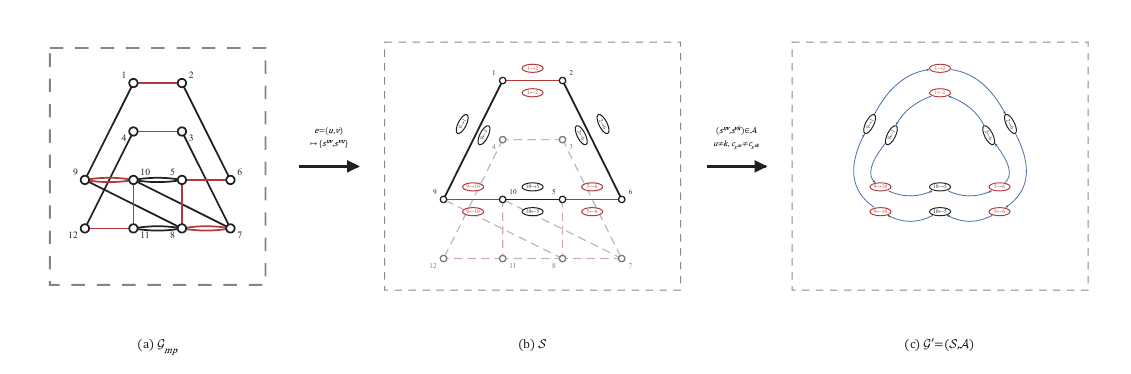}
\caption{Construction of auxiliary graph: (a) joint multigraph
$\mathcal{G}_{mp}$; (b) directional state nodes $s^{uv}$ and $s^{vu}$
associated with edge $(u,v)$; and (c) auxiliary directed graph
$\mathcal{G}'=(\mathcal{S},\mathcal{A})$.}
\label{fig:mpeax-construction}
\end{figure*}

Fig.~\ref{fig:mpeax-construction} illustrates this transformation. The edge group joining vertices 1 and 2 gives rise to two states, $s^{12}$ and $s^{21}$, representing the two traversal directions. Since $s^{12}$ ends at vertex 2, it can be followed by $s^{26}$ if the two groups belong to opposite sides of the joint multigraph. Likewise, $s^{21}$ can be followed by $s^{19}$ under the same condition.

\subsection{Budgeted heuristic algorithm for AB-cycles}
\label{sec:cycle-search}

After constructing the auxiliary graph $\mathcal{G}'$, mpEAX searches for directed cycles that induce legal \textit{AB-cycles}. A candidate cycle is a closed state sequence $\Gamma=(s_1,\ldots,s_\ell,s_{\ell+1})$, where $s_i\in\mathcal{S}$ and $s_{\ell+1}=s_1$. It is accepted if $(s_i,s_{i+1})\in\mathcal{A}$ for every $i=1,\ldots,\ell$, and the states $s_1,\ldots,s_\ell$ represent distinct undirected edge groups. The first condition enforces alternating traversal, while the second prevents repeated use of the same physical edge, including traversal in the opposite direction.

Enumerating all directed cycles in $\mathcal{G}'$ is computationally challenging when many parents are involved. We therefore use a budgeted heuristic algorithm (BHA) to generate a limited set of candidate directed cycles. BHA combines reverse breadth-first search (RBFS) with forward depth-first search (FDFS). For each selected start state node $s\in\mathcal{S}$, RBFS is performed on the reverse graph $\mathcal{G}'_R=(\mathcal{S},\mathcal{A}_R)$, where $\mathcal{A}_R=\{(s_i,s_j):(s_j,s_i)\in\mathcal{A}\}$, to compute the shortest return distance from each reachable state node to $s$. FDFS then uses these distances to prune partial paths that cannot return to $s$ or cannot close within the prescribed cycle-length limit.

\begin{figure}[!b]
\centering
\includegraphics[width=0.8\columnwidth]{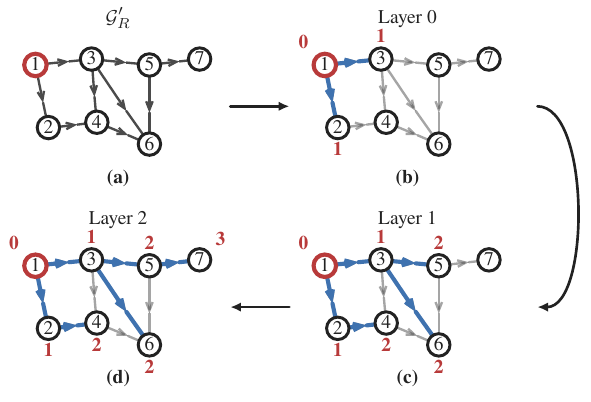}
\caption{Illustration of RBFS to generate distance labels.}
\label{fig:rbfs-layer-expansion}
\end{figure}

RBFS starts from a state node $s$ in the reverse graph $\mathcal{G}'_R$. Initially, all state nodes are unvisited. The algorithm marks $s$ as visited, sets $d_s(s)=0$, and places $s$ in a queue. It then repeatedly removes a state node $x$ from the queue and examines each of its out-neighbors $y$. If $y$ is unvisited, it is marked as visited, assigned $d_s(y)=d_s(x)+1$, and added to the queue. The procedure terminates when the queue is empty. Since each state node and arc is processed at most once, its time complexity is $O(|\mathcal{S}|+|\mathcal{A}|)$. Fig.~\ref{fig:rbfs-layer-expansion} illustrates the process with start state node 1. RBFS assigns $d_s(1)=0$ and then assigns distance 1 to state nodes 2 and 3. The subsequent layer expansions yield, for example, $d_s(7)=3$, meaning that state node 7 can return to $s$ in three transitions in the original graph $\mathcal{G}'$. States not reached by RBFS receive no finite return distance.

A path from $s$ to $x$ in the reverse graph $\mathcal{G}'_R$ corresponds to a path from $x$ back to $s$ in $\mathcal{G}'$. Thus, $d_s(x)$ gives the minimum number of transitions needed to return from $x$ to $s$. If RBFS does not reach $x$, no such return path exists, so $x$ cannot complete a cycle rooted at $s$.

Given a start node $s$ and the return-distance labels computed by RBFS, FDFS maintains its current path in a stack $C$, initially containing only $s$; the node at the top of $C$ is the current node. Only successors with finite return distances and edge groups not already in $C$ are eligible. These successors are ranked by whether their edge groups have been covered by previously accepted cycles, followed by increasing $d_s$ and out-degree; random values break any remaining ties. Inspired by the candidate-truncation strategy of beam search \cite{Lowerre1976}, FDFS considers at most the first $b$ successors in this order. A candidate successor $y$ is pruned if $|C|+d_s(y)>L_{\max}$, since even its shortest return path would exceed the cycle-length limit. If the top node can return directly to $s$, the path in $C$ closes a legal cycle.

After each FDFS attempt, BHA selects a new start node and repeats the search until the fixed search budget is exhausted or no eligible start nodes remain. The parameter $L_{\max}$ bounds the length of an accepted cycle, while $b$ limits the number of successors examined at each FDFS expansion. We denote these parameters by $\mathcal{B}=(L_{\max},b)$. Algorithms~\ref{alg:cycle-search} and~\ref{alg:fdfs} present the overall BHA procedure and its FDFS component, respectively.

\begin{algorithm}[!t]
\raggedright
\caption{Framework of the budgeted heuristic algorithm}
\label{alg:cycle-search}
\begin{algorithmic}[1]
\Require Auxiliary graph $\mathcal{G}'$, parameter vector $\mathcal{B}=(L_{\max},b)$.
\Ensure Set of legal directed cycles $\mathcal{C}$.
\State $\mathcal{G}'_R\gets\Call{ReverseGraph}{\mathcal{G}'}$
\State $\mathcal{C}\gets\emptyset$
\While{\Call{SearchBudgetAvailable}{}}
  \State $s\gets\Call{SelectStart}{\mathcal{G}',\mathcal{C}}$
  \State \textbf{if} $s=\emptyset$ \textbf{then break}
  \State $d_s\gets\Call{RBFS}{\mathcal{G}'_R,s}$
  \State $C\gets[s]$
  \Statex \hspace{\algorithmicindent}$\triangleright$ Initialize the FDFS path stack
  \State $\Gamma\gets\Call{FDFS}{\mathcal{G}',s,\mathcal{B},d_s,C,\mathcal{C}}$
  \State \textbf{if} $\Gamma\neq\emptyset$ \textbf{then} $\mathcal{C}\gets\mathcal{C}\cup\{\Gamma\}$
\EndWhile
\State \Return $\mathcal{C}$
\end{algorithmic}
\end{algorithm}

The \textsc{SelectStart} operation records previously tried start nodes, excludes edge groups covered by accepted cycles, and favors states with lower out-degree and greater capacity; it returns $\emptyset$ when no eligible start node remains.

Algorithm~\ref{alg:fdfs} details how FDFS extends the path stack $C$ initialized in Algorithm~\ref{alg:cycle-search}. At each call, the top of $C$ gives the current node. Eligible successors are ranked, and only the first $b$ are considered. FDFS pushes a successor before exploring it recursively and pops it when that branch fails. If the current node can return directly to $s$, the nodes in $C$, read from bottom to top and followed by $s$, form the returned closed cycle; this operation does not change the stack.

\begin{algorithm}[!t]
\raggedright
\caption{Forward depth-first search for a legal directed cycle}
\label{alg:fdfs}

\begingroup
\begin{algorithmic}[1]
\Require Auxiliary graph $\mathcal{G}'$, start state $s$, parameters $\mathcal{B}=(L_{\max},b)$, return-distance labels $d_s$, path stack $C$, and accepted-cycle set $\mathcal{C}$.
\Ensure A legal directed cycle $\Gamma$ or $\emptyset$.
\Function{FDFS}{$\mathcal{G}',s,\mathcal{B},d_s,C,\mathcal{C}$}
  \State $x\gets\Call{Top}{C}$
  \If{$(x,s)\in\mathcal{A}$}
    \State \Return $\Call{CloseCycle}{C,s}$
    \Statex \hspace{3em}$\triangleright$ Append $s$ to a copy of the path
  \EndIf
  \State $\mathcal{N}\gets\Call{Successors}{x}\setminus\{s\}$
  \State Remove unreachable states and states with edge groups already in $C$ from $\mathcal{N}$.

  \State $\mathcal{N}_b\gets\Call{SelectBeamCandidates}{\mathcal{N},b,\mathcal{C}}$
  \Statex \hspace{\algorithmicindent}$\triangleright$ Rank and retain the first $b$
  \ForAll{$y\in\mathcal{N}_b$}
    \If{$|C|+d_s(y)>L_{\max}$}
      \State \textbf{continue}
    \EndIf
    \State $\Call{Push}{C,y}$
    \State $\Gamma\gets\Call{FDFS}{\mathcal{G}',s,\mathcal{B},d_s,C,\mathcal{C}}$
    \If{$\Gamma\neq\emptyset$}
      \State \Return $\Gamma$
    \EndIf
    \State $\Call{Pop}{C}$
  \EndFor
  \State \Return $\emptyset$
\EndFunction
\end{algorithmic}
\endgroup
\end{algorithm}

Thus the search samples a tractable subset of the alternating-cycle space instead of decomposing the whole joint multigraph. Once an \textit{AB-cycle} is accepted, it is stored as the sequence of nodes, side indicators, and donor identifiers needed by the following \textit{E-set} construction.

\subsection{E-set construction and integration into the HGA}
\label{sec:eset-assembly}

After generating a set of \textit{AB-cycles}, mpEAX follows the main steps of HGA \cite{HeHaoWu2024}: assembling \textit{E-sets}, constructing intermediate offspring, and restoring exactly $p$ cycles. \textit{E-set} assembly requires an additional compatibility condition because distinct edge occurrences in $\mathcal{G}_{mp}$ may represent the same undirected edge. Consequently, two independently generated \textit{AB-cycles} may request the same edge operation when combined in one offspring.

Let $\mathcal{C}$ be the set of legal \textit{AB-cycles} obtained in Section~\ref{sec:cycle-search}. For a state $s^{uv}$, define $e(s^{uv})=\{u,v\}$ as its underlying undirected edge. For each $\Gamma\in\mathcal{C}$, let $F(\Gamma)=\{e(s):s\in\Gamma\}$ be the set of undirected edges represented by its states. A subset $\mathcal{C}_s\subseteq\mathcal{C}$ is admissible as one E-set only if
\begin{equation}
\begin{gathered}
F(\Gamma_i)\cap F(\Gamma_j)=\varnothing\
\text{for all distinct } \Gamma_i,\Gamma_j\in \mathcal{C}_s.
\end{gathered}
\label{eq:eset-compatible}
\end{equation}
This condition prevents repeated deletion or insertion of the same physical edge in one offspring. \textit{AB-cycles} in an \textit{E-set} may share vertices, but they cannot share an undirected edge.

The original HGA groups \textit{AB-cycles} that share vertices and, when necessary, merges these groups to obtain a bounded number of \textit{E-sets}. mpEAX follows the same procedure subject to Eq.~\eqref{eq:eset-compatible}. A cycle can be added to an \textit{E-set} only if its edge set $F(\Gamma)$ is disjoint from those of all cycles already included; likewise, two \textit{E-sets} can be merged only if their union remains admissible. At most $\beta$ \textit{E-sets} are retained, each producing one candidate offspring.

The resulting intermediate offspring are repaired using the 2-opt* procedure of HGA \cite{HeHaoWu2024} to restore exactly $p$ cycles. Replacing EAX with mpEAX in HGA yields the proposed algorithm {$\mathrm{HGA}_{mp}$}. A set of $k$ parents is chosen by repeated binary tournaments: two distinct population members are sampled, and the one with the lower objective value wins. A winner already selected for the current crossover is discarded, and selection continues until $k$ distinct parents have been obtained. The first selected parent is the base solution $\varphi_A$; the remaining $k-1$ parents form the donor set $\Phi_B$. All other components of HGA are retained.

\section{Experimental evaluation and comparisons}
\label{sec:experimental-evaluation}

In this section, we present computational experiments to evaluate the performance of {$\mathrm{HGA}_{mp}$} and compare it with state-of-the-art algorithms for the HpMP.

\subsection{Benchmark instances}

We conducted extensive experiments on four benchmark sets, $\mathbb{S}$, $\mathbb{M}$, $\mathbb{L}$, and $\mathbb{N}$, which have been used to evaluate algorithms for the HpMP. All instances are derived from TSP graphs in TSPLIB\footnote{\url{https://comopt.ifi.uni-heidelberg.de/software/TSPLIB95/}}. For each graph in sets $\mathbb{S}$, $\mathbb{M}$, and $\mathbb{L}$, previously used by Marzouk et~al.~\cite{marzouk2016branch} and {Herr{\'a}n et~al.~\cite{HerranColmenarDuarte2019}}, five instances are generated by setting $p$ to each value in $\left\{\left\lfloor n/10\right\rfloor,\left\lfloor n/7\right\rfloor,\left\lfloor n/5\right\rfloor,\left\lfloor n/4\right\rfloor,\left\lfloor n/3\right\rfloor\right\}$. For set $\mathbb{N}$, introduced by He et~al.~\cite{HeHaoWu2024}, seven instances are generated per graph, with $p\in\left\{\left\lfloor n/30\right\rfloor,\left\lfloor n/20\right\rfloor,\left\lfloor n/10\right\rfloor,\left\lfloor n/7\right\rfloor,\left\lfloor n/5\right\rfloor,\left\lfloor n/4\right\rfloor,\left\lfloor n/3\right\rfloor\right\}$. The resulting 215 instances and the solution certificates obtained by {$\mathrm{HGA}_{mp}$} are available online\footnote{The code for HGA {is} available at \url{https://github.com/pengfeihe-angers/HpMP-mp.git}.}.

\begin{itemize}
\item $\mathbb{S}$: 55 small instances derived from 11 TSPLIB graphs with 21 to 52 vertices.
\item$\mathbb{M}$: 55 medium instances derived from 11 TSPLIB graphs with 58 to 100 vertices.
\item  $\mathbb{L}$: 35 large instances derived from 7 TSPLIB graphs with 150 to 318 vertices.
\item $\mathbb{N}$: 70 very large instances derived from 10 TSPLIB graphs with 400 to 1060 vertices.
\end{itemize}

\subsection{Experimental protocol and reference algorithms}
\label{sec:protocol}

\noindent\textbf{Parameter setting.} The algorithmic components and parameters inherited from HGA follow the settings of He et~al.~\cite{HeHaoWu2024}. Table~\ref{tab:mpeax-parameters} lists only the parameters specific to mpEAX used in {$\mathrm{HGA}_{mp}$}. These settings are considered as the default parameter of the {$\mathrm{HGA}_{mp}$} algorithm used throughout the experiments.

\begin{table}[!t]
\centering
\caption{Parameters used in {$\mathrm{HGA}_{mp}$}.}
\label{tab:mpeax-parameters}
\begin{tabularx}{\columnwidth}{@{}lcXr@{}}
\hline
Parameter & Section & Description & Value \\
\hline
$L_{\max}$ & \ref{sec:cycle-search} & maximum length of an accepted \textit{AB-cycle} & 80 \\
$b$ & \ref{sec:cycle-search} & number of successors retained at each expansion &
8 \\
$k$ & \ref{sec:mpeax} & number of parent solutions used in mpEAX & 7\\
\hline
\end{tabularx}
\end{table}

\noindent\textbf{Reference algorithms.} In this paper, we use the best known solutions and two leading heuristic algorithms as references for evaluating the performance of {$\mathrm{HGA}_{mp}$}.

\begin{itemize}
\item \textbf{BKS.} The best-known solution value, which serves as an upper bound for each instance, as summarized from all state-of-the-art algorithms \cite{HeHaoWu2024,BarbatoGouveia2024,canas2025location}.

\item \textbf{PGVNS} {\cite{HerranColmenarDuarte2019}}. The algorithm was implemented in C++ and originally evaluated on a computer with a 2.5 GHz Intel Core i7-6500U processor and 8 GB of RAM. To facilitate a comparison with HGA, He et~al.~\cite{HeHaoWu2024} re-ran PGVNS on the platform used for their HGA experiments. We report those results under the heading ``re-PGVNS''.

\item \textbf{HGA} \cite{HeHaoWu2024}. The algorithm was implemented in C++, and its source code is available online\footnote{\url{https://github.com/pengfeihe-angers/HpMP.git}}. To facilitate a comparison with {$\mathrm{HGA}_{mp}$}, we re-run the code on the computer used in this paper and report the results under the heading ``re-HGA''.
\end{itemize}

\noindent\textbf{Experimental setting and stopping criterion.} The proposed algorithm was implemented in C++ and compiled using the \texttt{g++} compiler with the \texttt{-O3} option. All experiments were run on an AMD EPYC 7H12 processor (2.6~GHz and 2~GB of RAM) under Linux using a single thread. Each instance was independently executed 20 times with distinct random seeds. A run terminated after 500,000 iterations. We note that {$\mathrm{HGA}_{mp}$} and re-HGA are all run on the same computing platform to ensure a fair comparison among the algorithms.

\subsection{Computational results and comparisons}

\begin{table*}[!b]
\centering
\caption{Summary of results between {$\mathrm{HGA}_{mp}$} and the reference algorithms on the four sets of benchmark instances.}
\label{tab:mpeax-reference-comparison}
\normalsize
\begin{tabular*}{\textwidth}{@{\extracolsep{\fill}}llrrrrrrrr@{}}
\hline
\multirow{2}{*}{Set} & \multirow{2}{*}{Pair algorithms} &
\multicolumn{4}{c}{Best} & \multicolumn{4}{c}{Average} \\
\cmidrule(lr){3-6}\cmidrule(lr){7-10}
& & \#Wins & \#Ties & \#Losses & $p$-value & \#Wins & \#Ties & \#Losses & $p$-value \\
\midrule
\multirow{3}{*}{$\mathbb{S}$}
& {$\mathrm{HGA}_{mp}$} vs. BKS       & 0 & 55 & 0  & 1.00& -- & -- & -- & -- \\
& {$\mathrm{HGA}_{mp}$} vs. re-HGA    & 0 & 55 & 0  & 1.00& 0  & 55 & 0  & 1.00 \\
& {$\mathrm{HGA}_{mp}$} vs. re-PGVNS  & -- & -- & -- & -- & -- & -- & -- & -- \\
\hline
\multirow{3}{*}{$\mathbb{M}$}
& {$\mathrm{HGA}_{mp}$} vs. BKS       & 0 & 55 & 0 & 1.00 & -- & -- & -- & -- \\
& {$\mathrm{HGA}_{mp}$} vs. re-HGA    & 0 & 55 & 0  & 1.00 & 1  & 54 & 0 & 1.00 \\
& {$\mathrm{HGA}_{mp}$} vs. re-PGVNS  & 0 & 55 & 0  & 1.00& 6  & 49 & 0 & 0.031 \\
\hline
\multirow{3}{*}{$\mathbb{L}$}
& {$\mathrm{HGA}_{mp}$} vs. BKS       & 1 & 34 & 0 & 1.00 & -- & -- & -- & -- \\
& {$\mathrm{HGA}_{mp}$} vs. re-HGA    & 2 & 33 & 0 & 0.500 & 15 & 18 & 2 & 0.017 \\
& {$\mathrm{HGA}_{mp}$} vs. re-PGVNS  & 16 & 19 & 0 & $\ll 0.05$& 27 & 5 & 3 & $\ll 0.05$ \\
\hline
\multirow{3}{*}{$\mathbb{N}$}
& {$\mathrm{HGA}_{mp}$} vs. BKS       & 33 & 24 & 13 & 0.080 & -- & -- & -- & -- \\
& {$\mathrm{HGA}_{mp}$} vs. re-HGA    & 38 & 19 & 13 & 0.019 & 46 & 8 & 16 & 0.006 \\
& {$\mathrm{HGA}_{mp}$} vs. re-PGVNS  & 65 & 0 & 5 & $\ll 0.05$& 64 & 0 & 6 & $\ll 0.05$ \\
\hline
\end{tabular*}
\end{table*}

We report comparisons of {$\mathrm{HGA}_{mp}$} with the reference algorithms on the four sets of benchmark instances. Detailed computational results for each instance are presented in the Online Supplementary Material (Tables~S1--S4). The comparison summary is shown in Table~\ref{tab:mpeax-reference-comparison}, where the columns ``\#Wins'', ``\#Ties'', and ``\#Losses'' give the numbers of instances on which {$\mathrm{HGA}_{mp}$} obtains better, equal, or worse results, respectively. We use the Wilcoxon signed-rank test to assess pairwise performance differences between the compared algorithms at a significance level of $0.05$. A $p$-value below $0.05$ indicates a statistically significant difference. According to the summarized results of Table~\ref{tab:mpeax-reference-comparison}, the detailed results to be reported in the Online Supplementary Material, we make the following observations.

\begin{itemize}
\item {Sets $\mathbb{S}$ and $\mathbb{M}$.} For the small and medium instances, {$\mathrm{HGA}_{mp}$} and re-HGA perform almost identically. On set $\mathbb{S}$, both algorithms attain the BKS values and report the same best and average results on all 55 instances. On set $\mathbb{M}$, {$\mathrm{HGA}_{mp}$} matches the BKS values on all 55 instances. It ties re-HGA on the best results of all 55 instances and obtains one better average result. Compared with re-PGVNS, {$\mathrm{HGA}_{mp}$} ties the best results on all instances and obtains 6 better average results against no worse one; the small $p$-value ($0.031<0.05$) indicates that this difference is statistically significant. Thus, the solution quality of the compared algorithms is generally similar on these two sets.

\item {Set $\mathbb{L}$.} For the 35 large instances, {$\mathrm{HGA}_{mp}$} updates the BKS of one instance (new upper bound), matches 34 other BKS values, without missing any. Compared with re-HGA, {$\mathrm{HGA}_{mp}$} obtains 2 better and 33 equal best results without any worse one; for the average results, it obtains 15 better and 18 equal results, while the reverse is true for 2 instances. The difference is not statistically significant for the best results ($p>0.05$) but significant for the average results ($p<0.05$). The dominance of {$\mathrm{HGA}_{mp}$} over re-PGVNS is more evident: it obtains 16 better and 19 equal best results without any worse one, and 27 better and 5 equal average results against only 3 worse ones ($p<0.05$ for both comparisons).

\item {Set $\mathbb{N}$.} The advantage of {$\mathrm{HGA}_{mp}$} is more evident on the 70 very large-sized instances. {$\mathrm{HGA}_{mp}$} updates the BKS values of 33 instances (47\%), matches 24 other BKS values, and misses 13. Compared with re-HGA, {$\mathrm{HGA}_{mp}$} obtains 38 and 46 better results in terms of the best and average values, respectively, while the reverse is true for 13 and 16 cases; the small $p$-values indicate that both differences are statistically significant. The dominance of {$\mathrm{HGA}_{mp}$} over re-PGVNS is even more evident, with 65 better against 5 worse best results and 64 better against 6 worse average results ($p<0.05$ for both).
\end{itemize}

In summary, {$\mathrm{HGA}_{mp}$} and re-HGA provide similar results on {the small and medium sets $\mathbb{S}$ and $\mathbb{M}$}. {On the large set $\mathbb{L}$}, {$\mathrm{HGA}_{mp}$} matches re-HGA in terms of the best results and improves a number of average results. Its advantage is more pronounced on {the largest set $\mathbb{N}$}.

\FloatBarrier
\section{Additional experiments}
\label{sec:parent-effect}

This section examines the benefits and costs of increasing the parent count in mpEAX.

\subsection{Influences of the number of parent solutions in mpEAX}
\label{sec:parent-benefit}

\begin{figure*}[!b]
\centering
\includegraphics[width=0.8\textwidth]{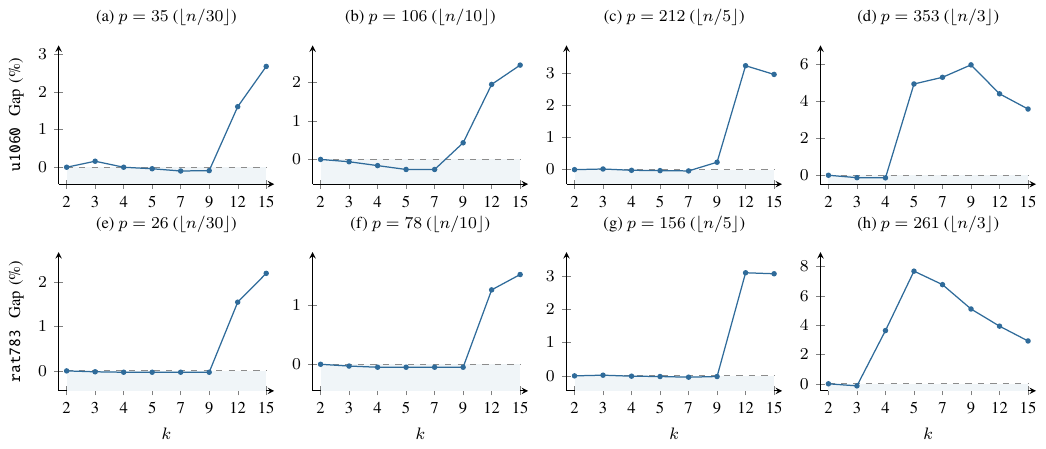}
\caption{mpEAX versus EAX on eight instances.}
\label{fig:best-gap-k}
\end{figure*}

To investigate the effect of the number of parents in mpEAX, we consider $k \in \{2,3,4,5,7,9,12,15\}$ and evaluate {$\mathrm{HGA}_{mp}$} on eight instances derived from \texttt{u1060} ($n=1060$) and \texttt{rat783} ($n=783$), each with $p \in \{\lfloor n/30\rfloor,\lfloor n/10\rfloor,\lfloor n/5\rfloor,\lfloor n/3\rfloor\}$. All experiments follow the protocol described in Section~\ref{sec:protocol}, with $k$ as the only parameter varied in {$\mathrm{HGA}_{mp}$}. Since {$\mathrm{HGA}_{mp}$} with $k=2$ reduces to re-HGA, we use re-HGA as the baseline. Fig.~\ref{fig:best-gap-k} reports the average gap in solution value for each value of $k$ relative to this baseline.

Except for the two instances shown in subfigures (d) and (h) of Fig.~\ref{fig:best-gap-k}, {$\mathrm{HGA}_{mp}$} exhibits a consistent performance pattern: as $k$ increases, the gap gradually decreases, reaches a minimum at an intermediate value of $k$, and then steadily increases. This pattern, observed on both graphs, indicates that a moderate number of parents improves performance, whereas further increases in $k$ erode these gains and eventually lead to deterioration. For the two instances with $p=\lfloor n/3\rfloor$, improvements over re-HGA occur only at the smallest values of $k$, while {$\mathrm{HGA}_{mp}$} underperforms this baseline for almost all larger values of $k$.

\begin{figure*}[t]
\centering
\includegraphics[width=0.8\textwidth]{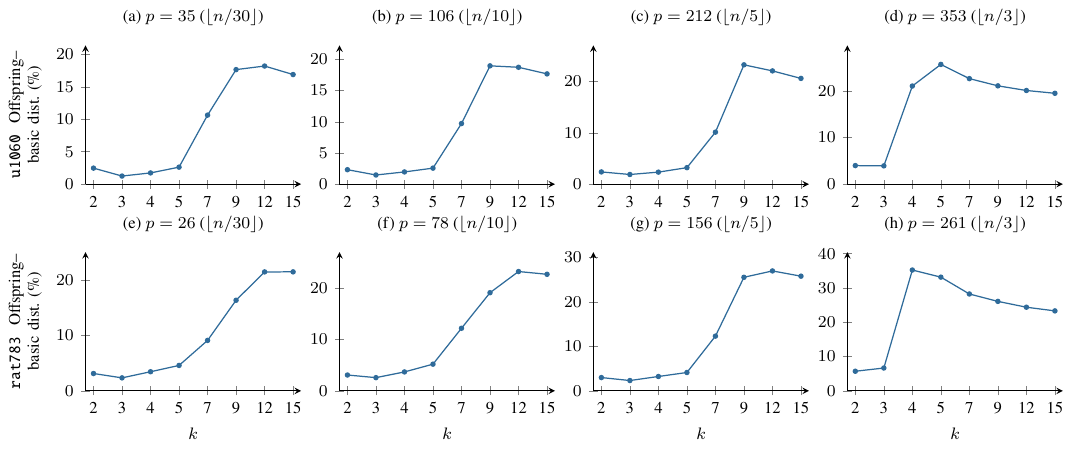}
\caption{The distance between offspring solutions and the base solution.}
\label{fig:offspring-basic-distance}
\end{figure*}

To help explain these performance differences, Fig.~\ref{fig:offspring-basic-distance} reports the mean normalized edge distance between offspring and the base solution, measured before local search. When $p<\lfloor n/3\rfloor$, this distance remains very small for $k=2$ and $k=3$ and then increases steadily with $k$, indicating that additional parents enable larger structural departures from the base solution. Comparing these results with Fig.~\ref{fig:best-gap-k} shows that the moderate displacement produced by intermediate values of $k$ coincides with smaller average solution-value gaps. This suggests that moderate structural variation benefits the search. As $k$ increases further, offspring deviate more substantially from the base solution, while the performance advantage diminishes and eventually reverses, suggesting that excessive structural disruption may offset the benefits of incorporating additional parents.

At $p=\lfloor n/3\rfloor$, the distance rises severalfold at $k=4$, where it reaches its maximum. Although it subsequently decreases gradually, it remains relatively high for all larger values of $k$. At this value of $p$, {$\mathrm{HGA}_{mp}$} underperforms re-HGA for almost every $k>2$, a pattern consistent with excessive structural disruption.

Even for large values of $k$, only a very small fraction of offspring edges fall outside the union of the selected parents’ edge sets. This indicates that departures from the base solution arise primarily through the recombination of parental edge structures \cite{NagataKobayashi2013}.

\begin{figure}[!htbp]
\centering
\includegraphics[width=\columnwidth]{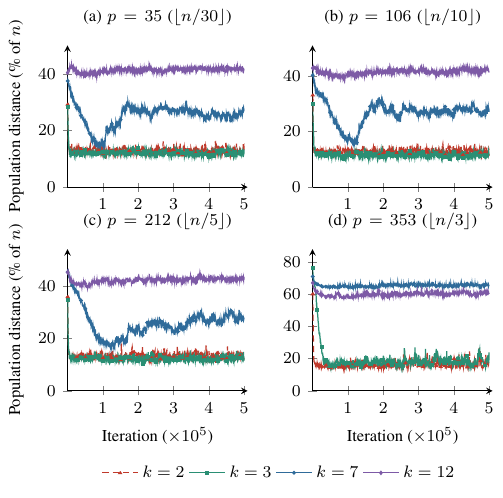}
\caption{Population distance over the search on \texttt{u1060}.}
\label{fig:population-distance}
\end{figure}

To further investigate the poorer performance of {$\mathrm{HGA}_{mp}$} at larger values of $k$ when $p=\lfloor n/3\rfloor$, we track population diversity using the mean pairwise edge distance among population members. A decline followed by stabilization at a low level indicates structural convergence, whereas a persistently high distance indicates sustained structural diversity. Fig.~\ref{fig:population-distance} plots this measure against the iteration count for $k\in\{2,3,7,12\}$ on the four instances derived from \texttt{u1060}, each corresponding to a different value of $p$.

In the first three subfigures of Fig.~\ref{fig:population-distance}, the mean pairwise edge distance remains low throughout the search for $k=2$ and $k=3$, indicating structurally cohesive populations. For $k=7$, the distance is initially high but decreases rapidly to a level close to that observed for $k=2$. It then rises and stabilizes at a moderate level, suggesting initial structural convergence followed by sustained, moderate diversity. By contrast, the distance for $k=12$ remains high throughout, indicating limited structural convergence. When $p=\lfloor n/3\rfloor$, the pattern changes: the initial decline for $k=7$ is absent, and the distance remains high throughout the search. Only $k=2$ and $k=3$ maintain structurally cohesive populations. These observations suggest that increasing $p$ reduces the number of parents that can be combined without compromising population convergence.

\subsection{\texorpdfstring{Why excessive parent counts degrade performance\\for large $p$}{Why excessive parent counts degrade performance for large p}}
\label{sec:parent-cost}

As shown in Table~S4 of the Online Supplementary Material, {$\mathrm{HGA}_{mp}$} underperforms re-HGA on 13 instances in set $\mathbb{N}$ when $p=\lfloor n/3\rfloor$. These instances have only $n/p\approx3$ vertices per route on average, so edge exchanges during crossover can disrupt several short routes simultaneously. Consistent with this interpretation, Fig.~\ref{fig:offspring-basic-distance} shows that, at this value of $p$, the distance between offspring and the base solution rises sharply as additional parents are introduced and remains high for larger values of $k$.

This pattern is also reflected in the population dynamics. Fig.~\ref{fig:population-distance} shows that, when $p=\lfloor n/3\rfloor$, the mean pairwise edge distance remains relatively low for $k=2$ and $k=3$ but stays high throughout the search for $k=7$ and $k=12$. Thus, populations with these larger parent counts exhibit limited structural convergence within the given computational budget. Such persistent dispersion may hinder the retention and reinforcement of useful edge structures through crossover and local search, consistent with the poorer performance at large $k$ reported in Fig.~\ref{fig:best-gap-k}.

\begin{figure}[!htbp]
% Keep Figures 7 and 8 together, preserving independent captions.
\centering
\includegraphics[width=\columnwidth]{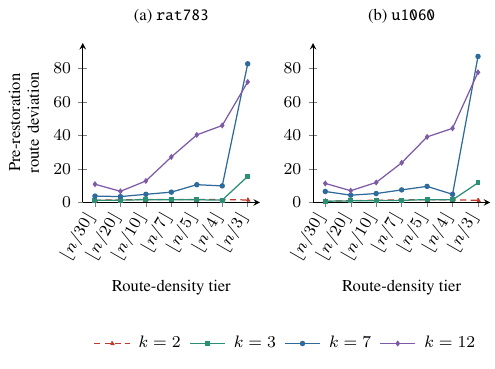}
\caption{Pre-restoration route-count deviation on \texttt{rat783} and
\texttt{u1060}.}
\label{fig:route-deviation-density}
\par\medskip
\includegraphics[width=\columnwidth]{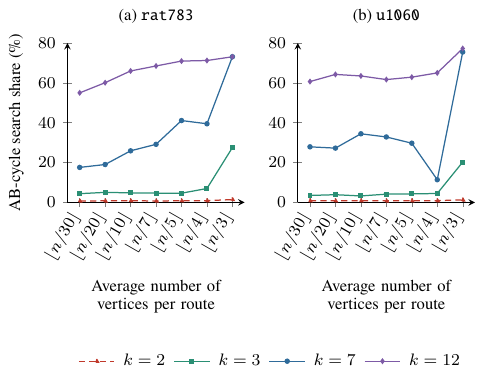}
\caption{Share of total runtime spent on AB-cycle search on \texttt{rat783}
and \texttt{u1060}.}
\label{fig:abcycle-search-share}
\end{figure}

Fig.~\ref{fig:route-deviation-density} provides further evidence of structural disruption: the route-count deviation before restoration is greatest when both $p$ and $k$ are large. Although restoration recovers feasibility, it may substantially modify the offspring structures produced by crossover. Fig.~\ref{fig:abcycle-search-share} further shows that the proportion of runtime devoted to \textit{AB-cycle} search increases with $k$, particularly at high route density. These results suggest that, while moderate parent counts provide useful diversity, excessive values of $k$ combine persistent population dispersion with greater crossover overhead, limiting the effectiveness of {$\mathrm{HGA}_{mp}$} on instances with large $p$.

\raggedbottom
\subsection[Sensitivity analysis]{Sensitivity analysis}
\label{sen_analysis}

We assess the sensitivity of {$\mathrm{HGA}_{mp}$} to the cycle-length limit $L_{\max}$ and candidate width $b$ by varying one parameter at a time. We vary $L_{\max}$ over $\{20,40,80,120,160\}$ while fixing $b=8$, and $b$ over $\{2,4,8,12,16\}$ while fixing $L_{\max}=80$. All other settings remain unchanged. The analysis uses the eight instances derived from \texttt{rat783} and \texttt{u1060} introduced in Section~\ref{sec:parent-benefit}, spanning four $p$ values. Each setting is evaluated in 20 independent runs with $k=7$ and a budget of 500,000 iterations per run.

\begin{figure}[!htbp]
\centering
\includegraphics[width=\columnwidth]{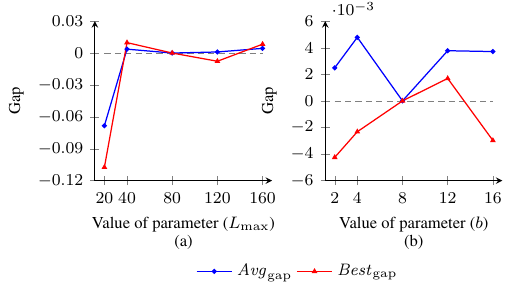}
\caption{Effects of $L_{\max}$ and $b$ on solution quality relative to the default configuration.}
\label{fig:search-parameter-sensitivity}
\end{figure}

For each parameter setting (PS), we compare the average and best solution values with those obtained under the default configuration $(L_{\max},b)=(80,8)$. The aggregate relative gaps across the eight selected instances are defined as
\begin{align}
\mathrm{Avg}_{\mathrm{gap}}
&= \sum_{i=1}^{8}
\frac{F_{\mathrm{PS},i}^{\mathrm{avg}}-F_{\mathrm{default},i}^{\mathrm{avg}}}
{F_{\mathrm{default},i}^{\mathrm{avg}}},
\label{eq:sensitivity-average-gap}\\
\mathrm{Best}_{\mathrm{gap}}
&= \sum_{i=1}^{8}
\frac{F_{\mathrm{PS},i}^{\mathrm{best}}-F_{\mathrm{default},i}^{\mathrm{best}}}
{F_{\mathrm{default},i}^{\mathrm{best}}}.
\label{eq:sensitivity-best-gap}
\end{align}
For instance $i$, $F_{\mathrm{PS},i}^{\mathrm{avg}}$ and $F_{\mathrm{PS},i}^{\mathrm{best}}$ denote the average and best solution values over 20 independent runs under PS, respectively; $F_{\mathrm{default},i}^{\mathrm{avg}}$ and $F_{\mathrm{default},i}^{\mathrm{best}}$ denote the corresponding values under the default configuration. Smaller gaps indicate better aggregate performance, with negative values favoring PS. Fig.~\ref{fig:search-parameter-sensitivity} presents the results.

Fig.~\ref{fig:search-parameter-sensitivity}(a) shows that $L_{\max}=20$ yields the lowest $\mathrm{Avg}_{\mathrm{gap}}$ and $\mathrm{Best}_{\mathrm{gap}}$ among the tested values. This aggregate advantage is concentrated on the two instances with $p=\lfloor n/3\rfloor$, where restricting cycle length improves both average and best solution values. On the other six instances, however, $L_{\max}=20$ produces worse average results than the default $L_{\max}=80$. These findings indicate that a shorter cycle-length limit is particularly beneficial at the highest route density but is not uniformly preferable. Increasing $L_{\max}$ beyond 80 provides no consistent benefit: relative to the default, $L_{\max}=120$ reduces $\mathrm{Best}_{\mathrm{gap}}$ but increases $\mathrm{Avg}_{\mathrm{gap}}$, whereas $L_{\max}=160$ increases both gaps.

Fig.~\ref{fig:search-parameter-sensitivity}(b) shows that the two measures also respond differently to candidate width. The default $b=8$ yields the lowest $\mathrm{Avg}_{\mathrm{gap}}$, whereas $b=2$ yields the lowest $\mathrm{Best}_{\mathrm{gap}}$. Larger candidate widths offer no consistent improvement. These results support $b=8$ when average solution quality is prioritized, but do not identify a single setting that performs best under both measures. The preferred cycle-length limit therefore depends on route density, while candidate width involves a trade-off between average and best performance.

\section{Conclusion}
\label{sec:conclusion}

In this paper, we propose a multi-parent edge assembly crossover, called mpEAX, for {the well-known} Hamiltonian \textit{p}-median problem. Unlike EAX, mpEAX combines one basic solution and multiple donor solutions in a single recombination operation. It uses a balanced residual multigraph, an auxiliary directed graph, and a budgeted \textit{AB-cycle} search to generate executable edge exchanges. Integrated into the existing HGA framework, mpEAX forms the new algorithm {$\mathrm{HGA}_{mp}$}.

Computational experiments on four sets of 215 benchmark instances show that {$\mathrm{HGA}_{mp}$} improves 34 best-known results and matches 168 others. Its advantage is particularly evident on the very large instances of set $\mathbb{N}$, where it improves 33 best-known results and significantly outperforms re-HGA in terms of both best and average solution values.

{Additional} experiments clarify the influence of the parent count. A moderate number of parents enlarges the exchange space and introduces useful diversity. However, when both $p$ and $k$ are large, the generated offspring become excessively dispersed, preventing the population from developing a shared high-quality structure and thereby weakening convergence. These findings indicate that the parent count should balance diversification and convergence. An interesting direction for future research is therefore to adapt $k$ dynamically according to the route density and the diversity of the current population. {This direction is related to dynamic-neighborhood search, where the search scale is adjusted during optimization \cite{Yao1992Dynamic}. In mpEAX, adapting the parent count could provide a means of regulating the recombination neighborhood to balance exploration and convergence.}

{In addition to the HpMP, mpEAX is also applicable to the TSP. Its joint multi-parent construction could also be extended to other EAX-based routing algorithms. For the minmax multiple traveling salesman problem, route segments from more parents could be recombined jointly while controlling disruption to the balance of tour lengths. For capacitated vehicle routing and its time-window variant, customer reassignment could be broadened while controlling potential increases in capacity and time-window violations. For latency location routing, sequential binary crossovers could be replaced by directed joint multi-parent edge assembly with the corresponding directed-edge balance. For split delivery vehicle routing, degree equalization would need to be extended to multiple parents and offspring delivery quantities coordinated. Overall, routing algorithms employing EAX have the potential to be extended to joint multi-parent crossover, but such extensions must be adapted to the solution representations, objectives, and constraints of each problem.}

\bibliographystyle{IEEEtran}
\bibliography{reference}
\end{document}